\documentclass{article}

\usepackage[preprint]{neurips_2019}

\usepackage[dvipsnames]{xcolor}

\usepackage[utf8]{inputenc} % allow utf-8 input
\usepackage[T1]{fontenc}    % use 8-bit T1 fonts
\usepackage{hyperref}       % hyperlinks
\usepackage{url}            % simple URL typesetting
\usepackage{booktabs}       % professional-quality tables
\usepackage{amsfonts}       % blackboard math symbols
\usepackage{nicefrac}       % compact symbols for 1/2, etc.
\usepackage{microtype}      % microtypography
\usepackage{import}
\usepackage{pdfpages}
\usepackage{float}

\usepackage{algorithm}
\usepackage{algorithmic}
\usepackage{wrapfig}
\usepackage{amsmath}
\DeclareMathOperator*{\argmax}{arg\,max}

\title{Uncertainty-aware Model-based Policy Optimization}

\makeatletter
\renewcommand{\author}[4][]{
      \stepcounter{custom@authors}
      \expandafter\def\csname custom@author\alph{custom@authors}\endcsname
      {\begin{tabular}[t]{c}\bf
      #2$^{\expandafter\the\csname custom@affiliationcounter#3\endcsname
        \if\relax\detokenize{#1}\relax\else,#1\fi}$ \\
        \texttt{#4}
       \end{tabular}}
}
\makeatother

\makeatletter
\def\thanks#1{\protected@xdef\@thanks{\@thanks
        \protect\footnotetext{#1}}}
\makeatother

\affiliation{VNU University of Engineering and Technology, Vietnam}{FIRSTAFF}
\affiliation{Microsoft Research, Redmond, USA}{THIRDAFF}

\author[]{Tung-Long Vuong}{FIRSTAFF}{}
\author[*]{Kenneth Tran\thanks{$^{*}$Corresponding author: ktran@microsoft.com}}{THIRDAFF}{}
\begin{document}
\maketitle

\begin{abstract}
Model-based reinforcement learning has the potential to be more sample efficient than model-free approaches. However, existing model-based methods are vulnerable to model bias, which leads to poor generalization and asymptotic performance compared to model-free counterparts. In addition, they are typically based on the model predictive control (MPC) framework, which not only is computationally inefficient at decision time but also does not enable policy transfer due to the lack of an explicit policy representation. In this paper, we propose a novel uncertainty-aware model-based policy optimization framework which solves those issues. In this framework, the agent simultaneously learns an uncertainty-aware dynamics model and optimizes the policy according to these learned models. In the optimization step, the policy gradient is computed by automatic differentiation through the models. With respect to sample efficiency alone, our approach shows promising results on challenging continuous control benchmarks with competitive asymptotic performance and significantly lower sample complexity than state-of-the-art baselines.
\end{abstract}

\section{Introduction} \label{sec:intro}
Reinforcement learning (RL) has been successfully applied to game playing. However, its application in real-world scenarios is still limited. One of the main challenges is the high sample complexity of the existing RL algorithms. The high sample complexity can be contributed by three factors: 
sample complexity of core learning algorithms, 
the ability of systems to learn from existing off-policy data, 
and the ability for a policy, as well as the world model in the case of model-based RL (MBRL), to be transferred to another task. 
We argue that in order for RL to be applicable in the real world, a RL system essentially needs to possess these properties: low sample complexity of the learning algorithm, the capability to learn from off-policy data, and transferability to similar tasks. 
To the best of our knowledge, none of the existing algorithms has all of those properties. 
In this paper, we propose a novel RL framework that is sample efficient, able to bootstrap from off-policy data, and also able to bootstrap from an existing policy.

Popular RL algorithms are divided into two main paradigms: model-free and model-based. 
Model-free algorithms have been shown to achieve good asymptotic performance in many high dimensional problems \cite{mnih2015human, silver2017mastering}. 
However, model-free ones have a crucial limitation that they often require a massive amount 
of data samples for training, thus prevents 
their applications from control/robotic domains, where the costs associated with real-system 
interactions are high. 
The main reason is that model-free RL learns the state/state-action values only from rewards and does not explicitly exploit the rich information underlying the transition dynamics data.

In contrast to model-free RL
(MFRL), model-based algorithms try to learn the transition dynamics, which is in turn used for imagining/planning without having to frequently interact with the real systems. Therefore, they are often considered more sample efficient than their model-free counterparts. 
In addition, the dynamics model is independent from rewards and thus can be transferred to other tasks in the same or similar environments. 
Nevertheless, the assumption about  the accuracy of the learned dynamics model is usually not satisfied, especially in complex environments. The model error and its
compounding effect when planning, i.e. a small bias in the model can lead to a highly erroneous value function estimate and a strongly-biased suboptimal policy, 
make MBRL less competitive in terms of asymptotic performance than MFRL for many non-trivial tasks. 

Many solutions have been proposed to mitigate the model bias issue. 
Early work such as \citet{deisenroth2011pilco} uses
Gaussian Processes (GP) to capture the model uncertainty. GP-based methods are, however, computationally intractable and unscalable for complex environments such as MuJoCo \cite{Todorov2012} or Atari games. In recent years, deep neural network (DNN) has been gaining lots of attention due to its high representational capacity and great successes in large-scale supervised training tasks. 
There have been several attempts in making DNN uncertainty-aware.  For examples, Bayesian neural networks were used in \citet{Gal2016, Depeweg2016, Kamthe2017}, 
or in \citet{gal2016dropout}, dropout was proposed as a scalable approximation to GP. 
More recently, bootstrapping or ensembling in general has become a favorable technique for uncertainty modeling in DNN such as in \citet{kurutach2018, clavera2018}. 
We hypothesize the reasons are because bootstrapping is a well-studied technique in Statistics and ensembling is relatively easy to train.

Another limitation of many existing MBRL 
methods is that they rely on the MPC framework. In this framework, at each time step $t$, the agent uses the learned dynamics model to plan $H$ (a hyperparameter) steps ahead, predicts the optimal sequence of actions $\{a_t, a_{t+1}, ..., a_{t+H}\}$, then interacts with the environment using $a_t$. 
But MPC, while easy to understand, has two drawbacks. First, each MPC step requires solving a high-dimensional optimization problem and thus is computationally prohibitive for applications requiring either real-time or low-latency reaction such as autonomous driving. Second, the policy is only implicit via solving the mentioned optimization problem. In more detail, not being able to explicitly represent the policy makes it hard to transfer the learned policy to other tasks or to initialize the agent with an existing better-than-random policy.

In contrast to MBRL, in the MFRL literature, policy gradient has shown to be an effective technique to train an agent \cite{lillicrap2015continuous, schulman2017proximal, haarnoja2018soft}. In this paper, we propose a new Model-based Policy Optimization (MBPO) framework that combines MBRL 
and policy gradient techniques in a principled way. Our experiments demonstrate the superior sample efficiency of our MBPO agent, compared to state-of-the-art methods, under the condition that the agent starts from scratch. 

\begin{figure}[H]
 \centering 
 \includegraphics[
    page=1,
    scale=0.7,
    keepaspectratio]{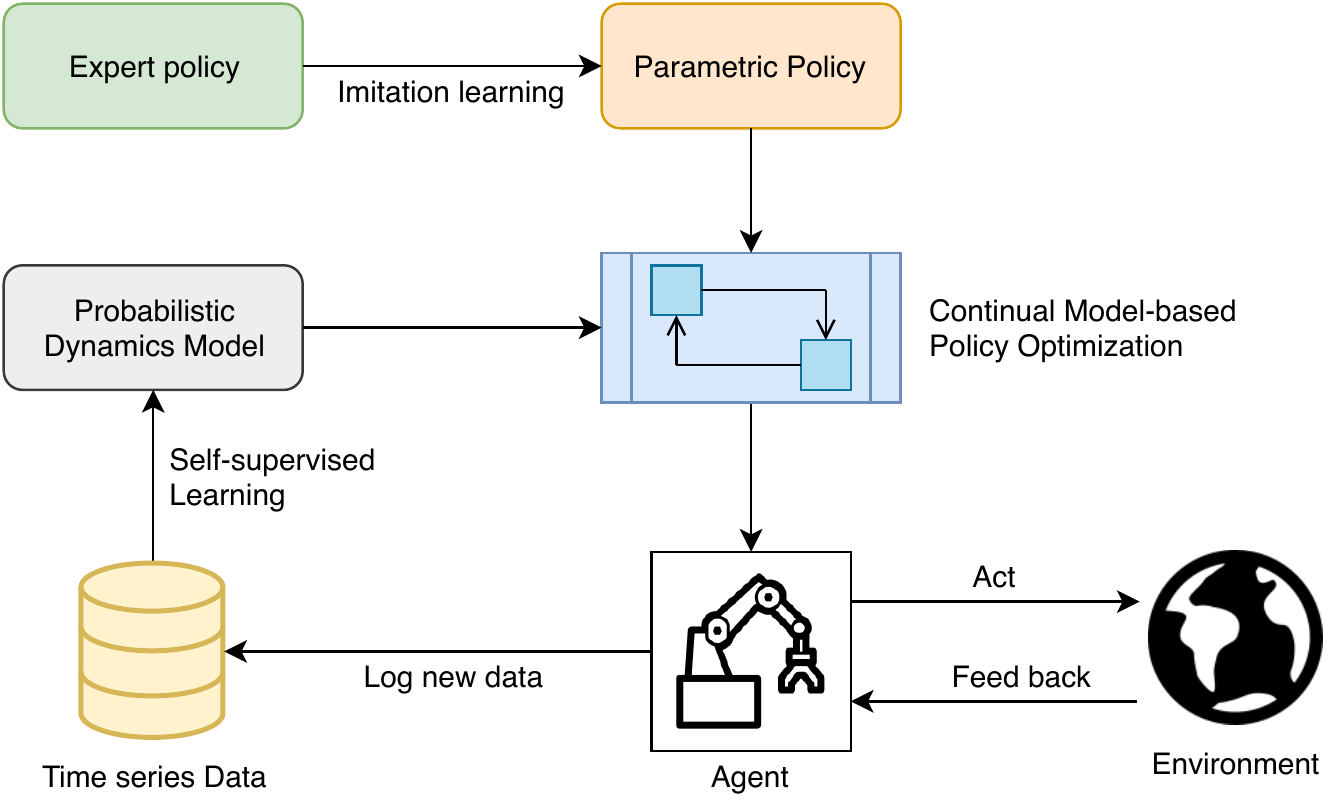}
 %\includepdf[pages=1]{}
 \label{fig:framework}
 \caption{Model-based Policy Optimization Framework}
\end{figure}

In addition to promising experimental results, we also emphasize that our framework is designed to support learning and improving from existing knowledge and from existing policy. Figure \ref{fig:framework} illustrates our broad approach to real-world reinforcement learning.

\section{Related work}
\label{sec:related}
Initial successes in MBRL in continuous control achieved promising results by learning control policies trained on models of local dynamics using linear parametric approximators \cite{abbeel2007, levine2013}. 
Alternate methods such as \citet{deisenroth2011pilco} incorporated non-parametric probabilistic GPs to capture model uncertainty during policy planning and evaluation. 
While these methods enhance data efficiency in low-dimensional tasks, their applications in more challenging domains such as environments involving non-contact dynamics and high-dimensional control remain limited by the inflexibility of their temporally local structure and intractable inference time. Our model, on the contrary, achieves both targets of having asymptotically high performance compared to MFRL methods and, at the same time, retaining data efficiency in those complex domains. 

Recently, there has been a revived interest in using DNNs to learn predictive models of environments from data, drawing inspiration from ideas in the early literature on this MBRL field. The large representational capacity of DNNs enables them as suitable function approximators for complex environments. However, additional care has to  be usually taken to avoid model bias, a situation where the DNNs overfit in the early stages of learning, resulting in inaccurate models. 
\citet{nagabandi2018} combined a learned dynamics network with MPC to initialize the policy network to accelerate learning in model-free deep RL. 
\citet{chua2018} extended this idea by introducing a bootstrapped ensemble of probabilistic DNNs to model predictive uncertainty of the learned networks and demonstrating that a pure model-based approach can attain the asymptotic performance of MFRL counterparts. However, the use of MPC to define a policy leads to poor run-time execution and hard to transfer policy across tasks.

Subsequent research proposed algorithms to leverage the learned ensemble of dynamics models to train a policy network. \citet{kurutach2018} learned 
a stochastic policy via trust-region policy optimization and \citet{clavera2018} casted the policy gradient as a meta-learning adaptation step with respect to each member of the ensemble. \citet{buckman2018} proposed an algorithm to learn a weighted combination of roll-outs of different horizon lengths, which dynamically interpolates between model-based and model-free learning based on the uncertainty in the model predictions. To our knowledge, this is the closest work in aside from ours, which learns a reward function in addition to the dynamics function. Furthermore, none of the aforementioned works propagates the uncertainty all the way to the value function and uses the concept of utility function to balance risk and return, 
as used in our model.

The ensemble of DNNs provide a straightforward technique to obtain reliable estimates of predictive uncertainty \cite{lakshminarayanan2017} and has 
been integrated with bootstrap to guide exploration in MFRL \cite{osband2016}. While many of the approaches mentioned in this section employ bootstrap to train an ensemble of models, we note that their implementations comprise of reconstructing bootstrap datasets at every training iteration, 
which effectively trains every single data sample and thus diminishes the advantage on uncertainty quantification achieved through bootstrap. Our model is different in that, to maintain online bootstrap datasets across the ensemble, it adds each incoming data sample to a dataset according to a Poisson probability distribution \cite{park2007, qin2013}, thereby guaranteeing asymptotically consistent bootstrap datasets.
\section{Uncertainty-aware Model-based Policy Optimization}
\label{sec:MBPO}
\subsection{Policy Optimization Formulation} \label{subsec:formulation}
Consider a discrete-time Markov Decision Process (MDP) defined by a tuple $M=\left \{S,A,F, R, H, \gamma \right \}$, in which $S$ is a state space, $A$ is an action space, $F:S\times A \rightarrow S$ is a deterministic transition function, $R:S\times A\rightarrow \mathbb{R}$ is a reward function, $T\in\mathbb{Z}_+$ is a task horizon, and $\gamma\in (0,1)$ is a discount factor. We define the return as the sum of rewards $R(s_t, a_t)$ along a trajectory $\tau := (s_{0}, a_{0}, ..., s_T, a_T)$ induced by a policy $\pi$ : $S \rightarrow  A$. The goal of reinforcement learning is to find a policy $\pi$ that maximizes the expected return, i.e.
\begin{equation}
    \argmax_\pi \mathbb{E}_{s_0}\left[V(\pi)(s_0) \right ],
\label{eqn:MDP}
\end{equation}
where $V(\pi)(s_0)=\sum_{t=0}^T \gamma^t R\left(s_t, \pi(s_t)\right)$, $s_{t} = F(s_{t-1}, \pi(s_{t-1}))$ for $t = 1, 2, ..., T$ and $s_0$ is randomly choosen from some initial dsitribution on $S$.

If the dynamics function $F$ and the reward function $R$ are given, solving \eqref{eqn:MDP} can be done using the Calculus of Variations \cite{young2000lecture} or Policy Gradient \cite{sutton2000policy}, which is the equivalent of Calculus of Variations when the control function is parameterized or is finite dimensional.

However, in reinforcement learning, $F$ and $R$ are often unknown and hence Equation \eqref{eqn:MDP} becomes a blackbox optimization problem with an unknown objective function. Following the Bayesian approach commonly used in the blackbox optimization literature \cite{shahriari2015taking}, we propose to solve this problem by iteratively learning a probabilistic estimate $\widehat{V}$ of $V$ from data and optimizing the policy according to this approximated model.

It is worth noting that any unbiased method would model $\widehat{V}(\pi)$ as a probabilistic estimate, i.e. $\widehat{V}(\pi)$ would be a distribution (as opposed to a point estimate) 
for a given $\pi$. 
Optimizing a stochastic objective is, however, not well-defined. 
Our solution is to transform $\widehat{V}$ into a deterministic utility function that reflects a subjective measure balancing the risk and return. Following \cite{markowitz1952portfolio, sato2001td, garcia2015comprehensive}, we propose a risk-sensitive objective criterion using a linear combination of the mean and the standard deviation of $\widehat{V}(\pi)$. Formally stated, our objective criterion now becomes 
\begin{equation}
    U(\pi)=\mathbb{E}_{s_0}\left[\mu\left(\widehat{V}(\pi)(s_0)\right) + c \times \sigma\left(\widehat{V}(\pi)(s_0)\right)\right],
\label{eqn:utility_objective}
\end{equation}
where $\mu$ and $\sigma$ are the mean and the standard deviation of $\widehat{V}(\pi)(s_0)$ respectively, and $c$ is a constant that represents the subjective risk preference of the learning agent. 
A positive risk preference infers that the agent is adventurous while a negative risk preference indicates that the agent has a safe 
exploration strategy. 

\subsection{Estimate of Value Function and Its Gradient} \label{subsec:estimation}
Section \ref{subsec:formulation} above 
provides a general framework for policy optimization under uncertainty, assuming the availability of the 
estimation model $\widehat{V}(\pi)$ of the 
true % added 
value function $V(\pi)$. 
In this section, we present a model-based method to compute $\widehat{V}(\pi)$ as an approximation of $V(\pi)$. 
The main idea is to approximate the functions $<F, R>$ with probabilistic parametric models $<\widehat{F}, \widehat{R}>$ and fully propagate the 
% estimation 
estimated 
uncertainty when planning under each policy $\pi$
% , 
from an initial state $s_0$. 
The value function estimate $\widehat{V}$ can be formulated as
\begin{equation}
    \widehat{V}(\pi)(s_0) = \sum_{t=0}^T \gamma^t \widehat{R}\left(\hat{s}_t, \pi(\hat{s}_t)\right),
    \label{eqn:rollout}
\end{equation}
where $\hat{s}_0=s_0$ and $\hat{s}_{t} = \widehat{F}(\hat{s}_{t-1}, \pi(\hat{s}_{t-1}))$ for $t=1, 2, ..., T$.
Next, we describe how to accurately model $<F, R>$ with well-calibrated uncertainty and a rollout technique that allows the uncertainty to be faithfully propagated into $\widehat{V}(\pi)$.

\subsubsection{Model Uncertainty with Online Bootstrap}
As discussed in Section 
\ref{sec:intro}, 
there are several prior attempts to learn uncertainty-aware dynamics models including GPs, Bayesian neural networks, dropout neural networks and ensemble of neural networks. In this work, however, we employ an ensemble of bootstrapped neural networks. Bootstrapping is a generic, principled 
and statistical approach for uncertainty quantification. Furthermore, as explained later in Section \ref{sec:backprop}, this ensemble approach also gives rise to easy gradient computation. In particular, $\widehat{F}$ is represented as $\left \{\widehat{f}_{\phi_{k}}(s_{t},a_{t})\rightarrow s_{t+1} \right \}_{k=1}^B$. 

For simplicity of implementation, we model each bootstrap replica as deterministic and rely on the ensemble as the sole mechanism for quantifying and propagating uncertainty. Each bootstrapped model $\widehat{f}_{\phi_{k}}$,
which is parameterized by $\phi_{k}$, learns to minimize the $l2$ one-step prediction loss over the respective bootstrapped dataset $\mathbb{D}_k$:

\begin{equation}
\min_{\phi_{k:=1 \mapsto B}}\mathbb{E}_{(s_t,a_t,s_{t+1})\sim  \mathbb{D}_k}\left \| s_{t+1} - \widehat{f}_{\phi_{k}} (s_{t},a_{t})\right \|^{2}.
\end{equation}

The training dataset $\mathbb{D}$ stores the transitions on which the agent has experienced. Since each model observes its own subset of the real data samples, the predictions across the ensemble remain sufficiently diverse in the early stages of the learning and 
will then converge to their true values as the error of the individual networks decreases. 

Bootstrap learning is often studied in the context of batch learning. However, since our agent updates its world model $\widehat{F}$ after each physical step for the best possible sample efficiency, we follow the online bootstrapping via sampling from Poisson distribution method presented in \citet{oza2005online, qin2013}. This is a very effective online approximation to batch bootstrapping, leveraging the following argument: bootstrapping a dataset $\mathbb{D}$ with $n$ examples means sampling $n$ examples from $\mathbb{D}$ with replacement. 
Each example $i$ will appear $z_i$ times in the bootstrapped sample where $z_i$ is a random variable whose distribution is $Binom(n, 1/n)$
because during resampling, the $i$-th example will have $n$ chances to be picked, each with probability $1/n$. This $Binom(n, 1/n)$ distribution converges to $Poisson(1)$ when $n\rightarrow\infty$. Therefore, for each new data point, this method adds $z_k$ copies of that data point to the bootstrapped dataset $\mathbb{D}_k$, where $z_k$ is sampled from a $Poisson(1)$.

Unlike many other model-based approaches, we also learn the reward function, along the same design of classical MBRL algorithms \cite{sutton1991dyna}. However, in our current implementation, we use a deterministic model for the reward function to simplify the policy evaluation. Note that unlike the error from the dynamics model, the error from the reward model does not get compounded when we estimate $\widehat{V}$.

\subsubsection{Bootstrap Rollout}
In this section, we describe how to propagate the estimates with uncertainty from the dynamics model to evaluate a policy $\pi$. We represent our policy $\pi_{\theta}$ : $S \rightarrow  A$ as a neural network parameterized by $\theta$ . Note that we choose to represent our policy as deterministic. We argue that while all estimation models, including that of the dynamics and of the value function, need to be stochastic (i.e. uncertainty aware), the policy does not need to be. The policy is not an estimator and deterministic policy simply means that the agent is consistent when taking an action, no matter how uncertain it may know about the world. 

Given a policy $\pi_\theta$ and an initial state $s_{0}\in\mathbb{D}$, we can estimate the distribution of $V(\pi)(s_0)$ by simulating $\pi$ through each each bootstrapped dynamics model. And since each bootstrap model is an independent approximator of the dynamics function, by expanding the value function via these dynamics approximators, we eventually obtain independent estimates  of that value function. Finally, these separate and independent trajectories collectively form an ensemble estimator of $V$.

In practice, we sample these trajectories with a finite horizon $H<T$. It is still a challenge to expand the value function estimation for a very long horizon due to a few reasons: 
\begin{itemize}
\item neural network training becomes harder when the depth increases,
\item despite our best effort to control the uncertainty, we still do not have a guarantee that our uncertainty modeling is perfectly calibrated, which in turn may be problematic if the planning horizon is too large, and 
\item the policy learning time is proportional to the rollout horizon.
\end{itemize}

\subsubsection{SGD and Gradient Computation}
\label{sec:backprop}
We can rewrite our objective as 
\begin{equation}
    \argmax_\theta J(\theta)=\mathbb{E}_{s}\left[ U_\theta(s)\right],
\label{eqn:rewrite_utility_objective}
\end{equation}
where $U_\theta(s)=\mu\left(\widehat{V}_{\theta}(s)\right) + c \times \sigma\left(\widehat{V}_{\theta}(s)\right)$. Using the ensemble method and the rollout technique described above, we can naturally compute $\mu\left(\widehat{V}_{\theta}(s)\right)$ and $\sigma\left(\widehat{V}_{\theta}(s)\right)$ for a given policy $\pi_\theta$ and for a given state $s$. Therefore, the policy $\pi_{\theta}$ can be updated using the SGD or a variant of it. 

The aforementioned rollout method also allows us to easily express $U(\theta)$ in 
Equation 
\eqref{eqn:rewrite_utility_objective} as a 
single % added 
computational graph of $\theta$. 
This makes it straight forward to compute the policy gradient $\nabla_\theta U(\theta)$ using automatic differentiation, a feature provided out-of-the-box in most popular deep learning toolkits.
\section{Algorithm Summary}
\begin{algorithm}[ht]
\caption{Uncertainty-aware Model-based Policy Optimization}
\label{alg:main}
\begin{algorithmic}[1]
\STATE Initialize bootstrapped datasets $\left \{ \mathbb{D}_i \right \}$, bootstrapped models $\widehat{F}=\left \{ \widehat{F}_i\right \}$, reward model $\widehat{R}$, and policy $\pi_\theta$.
\WHILE{not done}
  \STATE 1. Step in the environment, collect new data point $(s, a, {s}',r)$
  \STATE 2. Push data to the bootstrapped replay buffers: for each member $i$-th in the ensemble, add $z_i$ copies of that data point to $\left \{ \mathbb{D}_i \right \}$, where $z_i$ is sampled from $Poisson(1)$
  \STATE 3. Update  $\left \{  \widehat{F}_i\right \}_{i=1}^n$ and $\widehat{R}$
  \STATE 4. Compute the value of the policy $\widehat{V}^{H}_{\theta}(s)$ and the utility function $U_\theta(s)$ by simulating through the learned models $\{\widehat{F}_i\}$ and $\widehat{R}$.
  \STATE 5. Policy update using SGD with the policy gradient computed by backpropagating the gradient of $\mathbb{E}_{s} \left[ U(\theta)(s)\right ]$ through the models 
\ENDWHILE
\end{algorithmic}
\end{algorithm}
We summarize our method in Algorithm \ref{alg:main}. 
Furthermore, in this section, we also highlight some important details in our implementation.

\paragraph{Online off-policy learning.} Except for the initialization step (we may initialize the models with batch training from off-policy data), our model learning is an online learning process. For each time step, the learning cost stays constant and does not grow over time, which is required for lifelong learning. Despite being online, the learning is off-policy because we maintain a bootstrapped replay buffer for each model in the ensemble. For each model update, we sample a minibatch of training data from the respective replay buffer. In addition, as mentioned, the models can also be initialized from existing data even before the policy optimization starts.

\paragraph{Linearly weighted random sampling.}
Since our replay buffers are accumulated online, a naive uniformly sampling strategy would lead to early data being sampled more frequently than later data. 
We thus propose a linearly weighted random sampling scheme to mitigate the early-data bias issue. 
In this sampling scheme, example $i$-th is 
randomly sampled with weight $i$, i.e. higher weights for the fresher examples in each online update step. As proved in Appendix \ref{sec:linearly-weighted}, with this scheme, early data points are still cumulatively sampled slightly more frequently than later points but the cumulative bias gap is greatly reduced; and asymptotically all data points are equally sampled.
\section{Experiment}
\label{sec:experiments}
We provide two experimental analyses in this section. The first analyzes the model error when the planning horizon gets increased. 
The second provides the benchmark comparisons against state-of-the-art baselines.

\subsection{Model learning and compounding errors}
\begin{figure}[ht]
  \centering
  \includegraphics[width=1\textwidth]{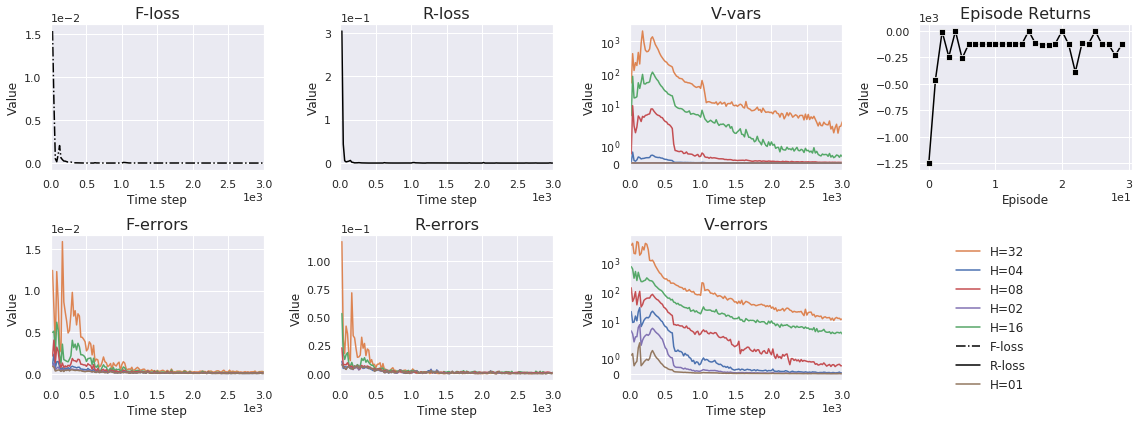}
  \caption{Our model prediction errors and compounding errors, both computed as mean square error, of value function estimate for different plan horizons $H$.}
  \label{fig:model_erros}
\end{figure}

Figure ~\ref{fig:model_erros} shows an error analysis of our model on the simple Pendulum environment. We used an oracle model to measure the errors. 
From the figure, we can see that even when the loss is small, the compounded error in the value function estimation can grow very quickly. Therefore, it is crucial to estimate the uncertainty (e.g. as variance or as error bound) in a principled way.

\subsection{Comparison to baseline algorithms}
We evaluate the performance of our proposed MBPO algorithm on three continuous control tasks in the MuJoCo simulator \cite{Todorov2012}: Pendulum-v0, Swimmer-v2 and HalfCheetah-v2 from OpenAI Gym \cite{brockman2016openai}.

\paragraph{Experimentation Protocol.}
We compare our algorithm against those state-of-the-art baseline algorithms designed for continuous control:
\begin{itemize}
  \item PPO \cite{schulman2017proximal}: a model-free policy gradient algorithm,
  \item DDPG \cite{lillicrap2015continuous}: an off-policy model-free actor-critic algorithm,
  \item SAC \cite{haarnoja2018soft}: a model-free actor-critic algorithm, which reports better data-efficiency than DDPG and PPO on most MuJoCo benchmarks,
  \item STEVE \cite{buckman2018}: a recent deterministic model-based algorithm.
\end{itemize}

Some other algorithms such as ME-TRPO \cite{kurutach2018}), MB-MPO \cite{clavera2018}, PETS \cite{chua2018} assume known reward function\footnote{We 
were 
also, with reasonable effort, unable to get their open-source implementations to run on standard OpenAI's Gym environments.}. We therefore do not include them in this benchmark study.

For each algorithm, we evaluate the learned policy after every episode (200 time steps for Pendulum-v0 and 1000 time steps for Swimmer-v2 and HalfCheetah-v2). 
The evaluation is done by running the current policy on 20 random episodes and then compute the average return over them. 

\begin{figure}[ht]
  \centering
  \includegraphics[width=1\textwidth]{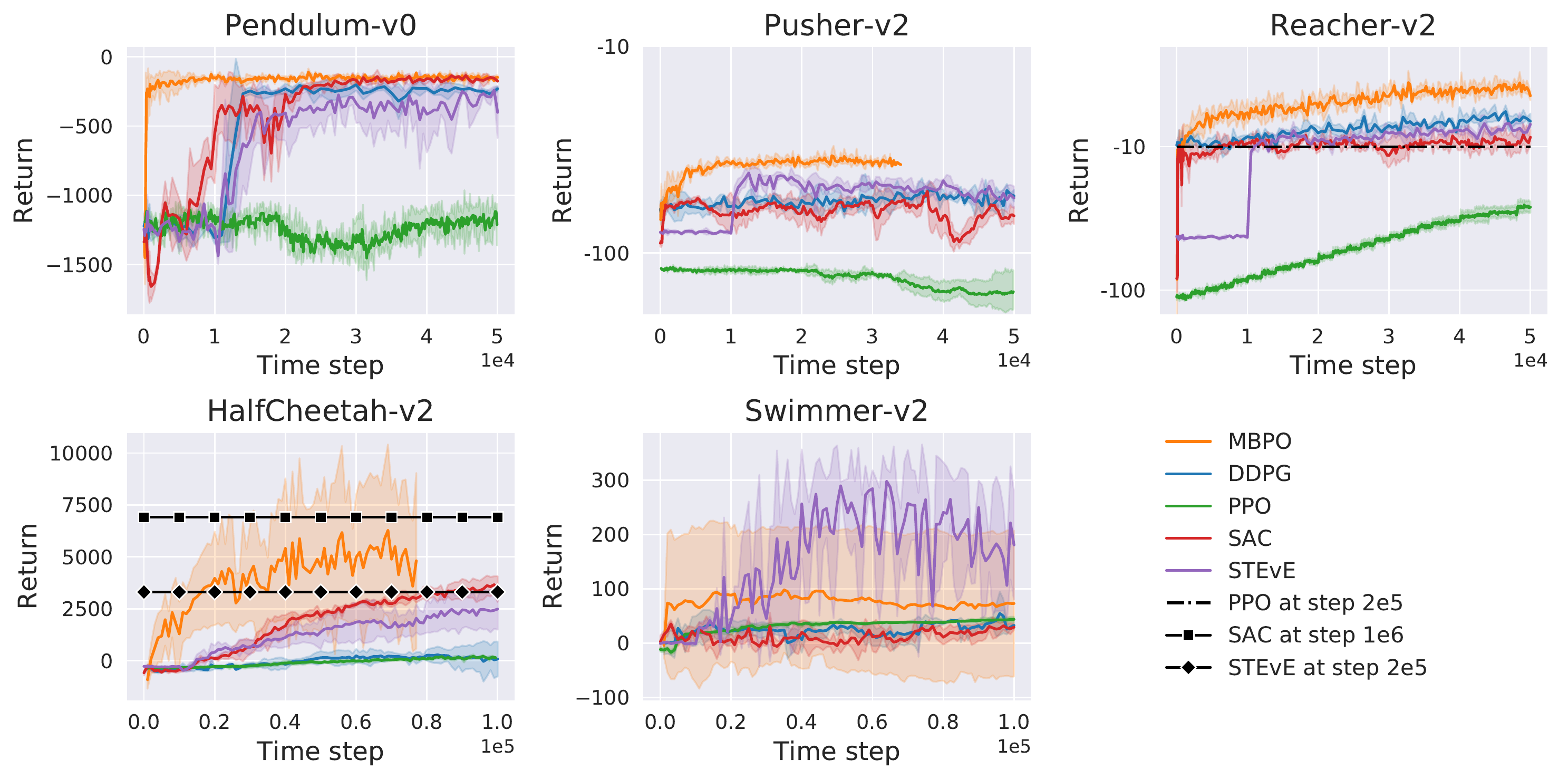}
  \caption{Average return of our MBPO model over 3 different randomly selected random seeds. Solid lines indicate the mean and shaded areas indicate one standard deviation.}
  \label{fig:average_perfromance}
\end{figure}

\begin{figure}[ht]
  \centering
  \includegraphics[width=1\textwidth]{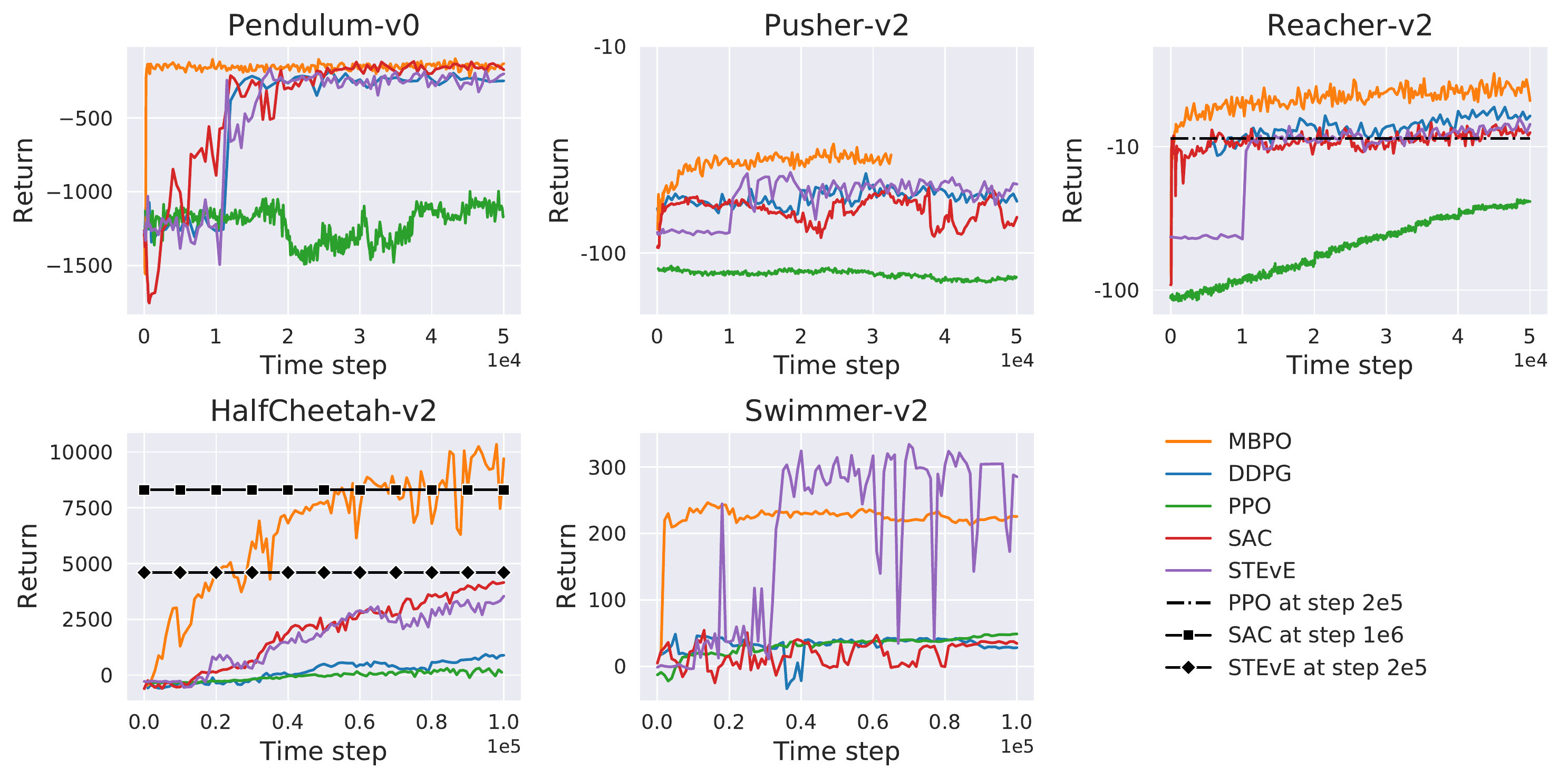}
  \caption{Best run of our MBPO model with different random seeds.}
  \label{fig:best_run}
\end{figure}

\paragraph{Results.} Figures \ref{fig:average_perfromance} and \ref{fig:best_run} show that MBPO has 
a superior sample efficiency compared to the baseline algorithms across a wide range of environments. 
Furthermore, it also has the asymptotic performance competitive to or even better than that of the model-free counterparts.

However, Figure \ref{fig:average_perfromance} also shows that the performance of MBPO is sensitive to the random seed. 
We hypothesize that this is due to our strategy of aggressive online learning and policy update after each step. It might also be due to our choice of learning early without random exploration on the first episode as many other methods apply. We plan to do a deeper analysis and address this instability issue in our future work.
% \section{Discussion}
\section{Discussion and Conclusion}
\label{sec:discussion}
Our experiments suggest that our MBPO algorithm not only can achieve the asymptotic performance of model-free methods in challenging continuous control tasks, it does so in much fewer samples. It is also more sample efficient than other existing MBRL algorithms. We further demonstrate that the model bias issue in model-based RL can be dealt with effectively with principled and careful uncertainty quantification. 

We acknowledge that our current implementation still has several limitations to overcome, such as high variance of the performance, which still depends on many hyper-parameters (plan horizon, risk sensitivity, and all hyper-parameters associated to neural network training) and even depends on the random seed. Note that these traits are not unique to our method. Nevertheless, the results indicate that if implemented right, model-based methods can be both sample efficient and has better asymptotic performance than model-free methods on challenging tasks.
In addition, by explicitly representing both the dynamics model and the policy, MBPO enables transfer learning, not just for the world (dynamics) model but also for the policy. 

Finally, we identify that sample efficiency, off-policy learning, and transferability are the three necessary, albeit not sufficient, properties for real-world reinforcement learning. % very strong and sensitive claim without much evidence in the paper, maybe be addressed in future work? 
We claim that our method meets these criteria and hence is a step towards real-world reinforcement learning. % another overly strong claim, maybe 

\bibliographystyle{plainnat}
\bibliography{citations.bib}

\begin{thebibliography}{31}
\providecommand{\natexlab}[1]{#1}
\providecommand{\url}[1]{\texttt{#1}}
\expandafter\ifx\csname urlstyle\endcsname\relax
  \providecommand{\doi}[1]{doi: #1}\else
  \providecommand{\doi}{doi: \begingroup \urlstyle{rm}\Url}\fi

\bibitem[Abbeel et~al.(2007)Abbeel, Coates, Quigley, and Ng]{abbeel2007}
Pieter Abbeel, Adam Coates, Morgan Quigley, and Andrew~Y Ng.
\newblock An application of reinforcement learning to aerobatic helicopter
  flight.
\newblock In \emph{Advances in neural information processing systems}, pages
  1--8, 2007.

\bibitem[Brockman et~al.(2016)Brockman, Cheung, Pettersson, Schneider,
  Schulman, Tang, and Zaremba]{brockman2016openai}
Greg Brockman, Vicki Cheung, Ludwig Pettersson, Jonas Schneider, John Schulman,
  Jie Tang, and Wojciech Zaremba.
\newblock Openai gym.
\newblock \emph{arXiv preprint arXiv:1606.01540}, 2016.

\bibitem[Buckman et~al.(2018)Buckman, Hafner, Tucker, Brevdo, and
  Lee]{buckman2018}
Jacob Buckman, Danijar Hafner, George Tucker, Eugene Brevdo, and Honglak Lee.
\newblock Sample-efficient reinforcement learning with stochastic ensemble
  value expansion.
\newblock In \emph{Advances in Neural Information Processing Systems}, pages
  8224--8234, 2018.

\bibitem[Chua et~al.(2018)Chua, Calandra, McAllister, and Levine]{chua2018}
Kurtland Chua, Roberto Calandra, Rowan McAllister, and Sergey Levine.
\newblock Deep reinforcement learning in a handful of trials using
  probabilistic dynamics models.
\newblock In \emph{Advances in Neural Information Processing Systems}, pages
  4754--4765, 2018.

\bibitem[Clavera et~al.(2018)Clavera, Rothfuss, Schulman, Fujita, Asfour, and
  Abbeel]{clavera2018}
Ignasi Clavera, Jonas Rothfuss, John Schulman, Yasuhiro Fujita, Tamim Asfour,
  and Pieter Abbeel.
\newblock Model-based reinforcement learning via meta-policy optimization.
\newblock \emph{arXiv preprint arXiv:1809.05214}, 2018.

\bibitem[Deisenroth and Rasmussen(2011)]{deisenroth2011pilco}
Marc Deisenroth and Carl~E Rasmussen.
\newblock Pilco: A model-based and data-efficient approach to policy search.
\newblock In \emph{Proceedings of the 28th International Conference on machine
  learning (ICML-11)}, pages 465--472, 2011.

\bibitem[Depeweg et~al.(2016)Depeweg, Hern{\'a}ndez-Lobato, Doshi-Velez, and
  Udluft]{Depeweg2016}
Stefan Depeweg, Jos{\'e}~Miguel Hern{\'a}ndez-Lobato, Finale Doshi-Velez, and
  Steffen Udluft.
\newblock Learning and policy search in stochastic dynamical systems with
  bayesian neural networks.
\newblock \emph{arXiv preprint arXiv:1605.07127}, 2016.

\bibitem[Gal and Ghahramani(2016)]{gal2016dropout}
Yarin Gal and Zoubin Ghahramani.
\newblock Dropout as a bayesian approximation: Representing model uncertainty
  in deep learning.
\newblock In \emph{international conference on machine learning}, pages
  1050--1059, 2016.

\bibitem[Gal et~al.(2016)Gal, McAllister, and Rasmussen]{Gal2016}
Yarin Gal, Rowan McAllister, and Carl~Edward Rasmussen.
\newblock Improving pilco with bayesian neural network dynamics models.
\newblock In \emph{Data-Efficient Machine Learning workshop, ICML}, volume~4,
  2016.

\bibitem[Garc{\i}a and Fern{\'a}ndez(2015)]{garcia2015comprehensive}
Javier Garc{\i}a and Fernando Fern{\'a}ndez.
\newblock A comprehensive survey on safe reinforcement learning.
\newblock \emph{Journal of Machine Learning Research}, 16\penalty0
  (1):\penalty0 1437--1480, 2015.

\bibitem[Haarnoja et~al.(2018)Haarnoja, Zhou, Abbeel, and
  Levine]{haarnoja2018soft}
Tuomas Haarnoja, Aurick Zhou, Pieter Abbeel, and Sergey Levine.
\newblock Soft actor-critic: Off-policy maximum entropy deep reinforcement
  learning with a stochastic actor.
\newblock \emph{arXiv preprint arXiv:1801.01290}, 2018.

\bibitem[Kamthe and Deisenroth(2017)]{Kamthe2017}
Sanket Kamthe and Marc~Peter Deisenroth.
\newblock Data-efficient reinforcement learning with probabilistic model
  predictive control.
\newblock \emph{arXiv preprint arXiv:1706.06491}, 2017.

\bibitem[Kurutach et~al.(2018)Kurutach, Clavera, Duan, Tamar, and
  Abbeel]{kurutach2018}
Thanard Kurutach, Ignasi Clavera, Yan Duan, Aviv Tamar, and Pieter Abbeel.
\newblock Model-ensemble trust-region policy optimization.
\newblock \emph{arXiv preprint arXiv:1802.10592}, 2018.

\bibitem[Lakshminarayanan et~al.(2017)Lakshminarayanan, Pritzel, and
  Blundell]{lakshminarayanan2017}
Balaji Lakshminarayanan, Alexander Pritzel, and Charles Blundell.
\newblock Simple and scalable predictive uncertainty estimation using deep
  ensembles.
\newblock In \emph{Advances in Neural Information Processing Systems}, pages
  6402--6413, 2017.

\bibitem[Levine and Koltun(2013)]{levine2013}
Sergey Levine and Vladlen Koltun.
\newblock Guided policy search.
\newblock In \emph{International Conference on Machine Learning}, pages 1--9,
  2013.

\bibitem[Lillicrap et~al.(2015)Lillicrap, Hunt, Pritzel, Heess, Erez, Tassa,
  Silver, and Wierstra]{lillicrap2015continuous}
Timothy~P Lillicrap, Jonathan~J Hunt, Alexander Pritzel, Nicolas Heess, Tom
  Erez, Yuval Tassa, David Silver, and Daan Wierstra.
\newblock Continuous control with deep reinforcement learning.
\newblock \emph{arXiv preprint arXiv:1509.02971}, 2015.

\bibitem[Markowitz(1952)]{markowitz1952portfolio}
Harry Markowitz.
\newblock Portfolio selection.
\newblock \emph{The journal of finance}, 7\penalty0 (1):\penalty0 77--91, 1952.

\bibitem[Mnih et~al.(2015)Mnih, Kavukcuoglu, Silver, Rusu, Veness, Bellemare,
  Graves, Riedmiller, Fidjeland, Ostrovski, et~al.]{mnih2015human}
Volodymyr Mnih, Koray Kavukcuoglu, David Silver, Andrei~A Rusu, Joel Veness,
  Marc~G Bellemare, Alex Graves, Martin Riedmiller, Andreas~K Fidjeland, Georg
  Ostrovski, et~al.
\newblock Human-level control through deep reinforcement learning.
\newblock \emph{Nature}, 518\penalty0 (7540):\penalty0 529, 2015.

\bibitem[Nagabandi et~al.(2017)Nagabandi, Kahn, Fearing, and
  Levine]{nagabandi2018}
Anusha Nagabandi, Gregory Kahn, Ronald~S. Fearing, and Sergey Levine.
\newblock Neural network dynamics for model-based deep reinforcement learning
  with model-free fine-tuning.
\newblock \emph{CoRR}, abs/1708.02596, 2017.
\newblock URL \url{http://arxiv.org/abs/1708.02596}.

\bibitem[Osband et~al.(2016)Osband, Blundell, Pritzel, and Van~Roy]{osband2016}
Ian Osband, Charles Blundell, Alexander Pritzel, and Benjamin Van~Roy.
\newblock Deep exploration via bootstrapped dqn.
\newblock In \emph{Advances in neural information processing systems}, pages
  4026--4034, 2016.

\bibitem[Oza(2005)]{oza2005online}
Nikunj~C Oza.
\newblock Online bagging and boosting.
\newblock In \emph{2005 IEEE international conference on systems, man and
  cybernetics}, volume~3, pages 2340--2345. Ieee, 2005.

\bibitem[Park et~al.(2007)Park, Ostrouchov, and Samatova]{park2007}
Byung-Hoon Park, George Ostrouchov, and Nagiza~F Samatova.
\newblock Sampling streaming data with replacement.
\newblock \emph{Computational Statistics \& Data Analysis}, 52\penalty0
  (2):\penalty0 750--762, 2007.

\bibitem[Qin et~al.(2013)Qin, Petricek, Karampatziakis, Li, and
  Langford]{qin2013}
Zhen Qin, Vaclav Petricek, Nikos Karampatziakis, Lihong Li, and John Langford.
\newblock Efficient online bootstrapping for large scale learning.
\newblock \emph{arXiv preprint arXiv:1312.5021}, 2013.

\bibitem[Sato et~al.(2001)Sato, Kimura, and Kobayashi]{sato2001td}
Makoto Sato, Hajime Kimura, and Shibenobu Kobayashi.
\newblock Td algorithm for the variance of return and mean-variance
  reinforcement learning.
\newblock \emph{Transactions of the Japanese Society for Artificial
  Intelligence}, 16\penalty0 (3):\penalty0 353--362, 2001.

\bibitem[Schulman et~al.(2017)Schulman, Wolski, Dhariwal, Radford, and
  Klimov]{schulman2017proximal}
John Schulman, Filip Wolski, Prafulla Dhariwal, Alec Radford, and Oleg Klimov.
\newblock Proximal policy optimization algorithms.
\newblock \emph{arXiv preprint arXiv:1707.06347}, 2017.

\bibitem[Shahriari et~al.(2015)Shahriari, Swersky, Wang, Adams, and
  De~Freitas]{shahriari2015taking}
Bobak Shahriari, Kevin Swersky, Ziyu Wang, Ryan~P Adams, and Nando De~Freitas.
\newblock Taking the human out of the loop: A review of bayesian optimization.
\newblock \emph{Proceedings of the IEEE}, 104\penalty0 (1):\penalty0 148--175,
  2015.

\bibitem[Silver et~al.(2017)Silver, Schrittwieser, Simonyan, Antonoglou, Huang,
  Guez, Hubert, Baker, Lai, Bolton, et~al.]{silver2017mastering}
David Silver, Julian Schrittwieser, Karen Simonyan, Ioannis Antonoglou, Aja
  Huang, Arthur Guez, Thomas Hubert, Lucas Baker, Matthew Lai, Adrian Bolton,
  et~al.
\newblock Mastering the game of go without human knowledge.
\newblock \emph{Nature}, 550\penalty0 (7676):\penalty0 354, 2017.

\bibitem[Sutton(1991)]{sutton1991dyna}
Richard~S Sutton.
\newblock Dyna, an integrated architecture for learning, planning, and
  reacting.
\newblock \emph{ACM SIGART Bulletin}, 2\penalty0 (4):\penalty0 160--163, 1991.

\bibitem[Sutton et~al.(2000)Sutton, McAllester, Singh, and
  Mansour]{sutton2000policy}
Richard~S Sutton, David~A McAllester, Satinder~P Singh, and Yishay Mansour.
\newblock Policy gradient methods for reinforcement learning with function
  approximation.
\newblock In \emph{Advances in neural information processing systems}, pages
  1057--1063, 2000.

\bibitem[Todorov et~al.(2012)Todorov, Erez, and Tassa]{Todorov2012}
Emanuel Todorov, Tom Erez, and Yuval Tassa.
\newblock Mujoco: A physics engine for model-based control.
\newblock In \emph{2012 IEEE/RSJ International Conference on Intelligent Robots
  and Systems}, pages 5026--5033. IEEE, 2012.

\bibitem[Young(2000)]{young2000lecture}
Laurence~Chisholm Young.
\newblock \emph{Lecture on the calculus of variations and optimal control
  theory}, volume 304.
\newblock American Mathematical Soc., 2000.

\end{thebibliography}
\appendix
\section{Appendix}
\subsection{Why linearly weighted random sampling is a fairer sampling scheme}
\label{sec:linearly-weighted}
Consider the following online learning process: for each time step, we need to randomly sample an example from the accumulating dataset. Suppose that at time $t$, each example $i$-th is randomly sampled with weight $w(t,i)$. Note that at each time $t$, we have a total of $t$ examples in the dataset. Then the probability of that example being sampled is 
\[
\frac{w(t,i)}{\sum_{k=1}^t w(t,k)}.
\]
If we use uniformly random sampling then the expected number of times an example $i$-th gets selected until time $t$ is 
\[
C_i ^t= \sum_{k=i}^t\frac{1}{k}.
\]
Hence, for all $t$, for $i>j$, $C^t_i$ is larger than $C^t_j$ by $\sum_{k=i}^j \frac{1}{k}$.
Now, if we use a linearly weighted random sampling scheme, in which $w(t,i)=i$, then the expected number of times an example $i$-th gets selected until time $t$ is 
\[
C_i ^t = \sum_{k=i}^t\frac{2i}{k(k+1)} =2\sum_{k=i}^t\left(\frac{i}{k}-\frac{i}{k+1}\right) =2 - 2\frac{i}{t}.
\]
% Now, 
We can see that 
at time $t$, $C_i^t$ is still larger than $C_j^t$ for $i<j$ but by weighting recent examples more in each online update step, we reduce the overall early-data bias. 
    
\subsection{Environments}

We evaluate the performance of our proposed MBPO algorithm on five continuous control tasks in the MuJoCo simulator from OpenAI Gym and we keep the default configurations prodived by OpenAI Gym.

\begin{table}[ht]
  \caption{Description of the environment used for testing}
  \label{table:nn_configs}
  \centering
  \begin{tabular}{lccc}
    \toprule
     Environment & State dimension & Action dimension & Task horizon \\
    \midrule
    Reacher-v2 & $11$  & $2$ & $50$    \\
    Pusher-v2 & $23$   & $27$ & $100$ \\
    Pendulum-v0 & $3$  & $1$ & $200$    \\
    Swimmer-v2 & $8$  & $2$ & $1000$    \\
    HalfCheetah-v2 & $23$  & $6$ & $1000$  \\
    \bottomrule
  \end{tabular}
\end{table}
% \subsection{Network hyper-parameters setting}

% \begin{table}[ht]
%   \caption{Network layer configurations}
%   \label{table:nn_configs}
%   \centering
%   \begin{tabular}{llll}
%     \toprule
%      Env & F & R & Policy \\
%     \midrule
%     Pendulum-v0 & $[4, 64, 64, 3]$  & $[4, 64, 64, 1]$ & $[3, 8, 1]$    \\
%     HalfCheetah-v2 & $[23, 1024, 512, 256, 17]$  & $[23, 1024, 512, 256, 1]$ & $[17, 512, 256, 6]$    \\
%     Swimmer-v2 & $[10, 256, 128, 8]$  & $[10, 256, 128, 1]$ & $[8, 128, 128, 2]$    \\
%     Pusher-v2 & $[30, 1024, 512, 256, 23]$  & $[30, 1024, 512, 256, 1]$ & $[23, 512, 256, 7]$    \\
%     Reacher-v2 & $[13, 256, 128, 11]$  & $[13, 256, 128, 1]$ & $[11, 128, 128, 2]$    \\
    
%     \bottomrule
%   \end{tabular}
% \end{table}

\end{document}